\documentclass[twocolumn]{asme2e}

\newcommand{\be}{\begin{equation}}
\newcommand{\ee}{\end{equation}}
\newcommand{\beq}{\begin{equation}}
\newcommand{\eeq}{\end{equation}}
\newcommand{\bed}{\begin{displaymath}}
\newcommand{\eed}{\end{displaymath}}
\newcommand{\beqa}{\begin{eqnarray}}
\newcommand{\eeqa}{\end{eqnarray}}
\newcommand{\beqann}{\begin{eqnarray*}}
\newcommand{\eeqann}{\end{eqnarray*}}
\newcommand{\bseq}{\begin{subequations}}
\newcommand{\eseq}{\end{subequations}}

\newcommand{\ba}{\begin{array}}
\newcommand{\ea}{\end{array}}

\newcommand{\negr}[1]{{\bf {#1}}}

\usepackage{amsbsy}
\usepackage{subeqn}
\usepackage[dvips]{graphicx}
\def\confshortname{DETC'2002}
\def\conffullname{2002 ASME Design Engineering Technical Conferences}
\def\confdate{29}
\def\confmonth{September-October}
\def\confyear{2002}
\def\confcity{Montreal}
\def\confstate{Quebec}
\def\confcountry{Canada}

\def\papernum{DETC2002/MECH-34268}

\title{The Isoconditioning Loci of Planar Three-DOF Parallel Manipulators }

\author{D. Chablat, Ph. Wenger, S. Caro
    \affiliation{
      Institut de Recherche en \\ Communications et Cybern\'etique de Nantes
      \thanks{IRCCyN: UMR n$^\circ$ 6597 CNRS, \'Ecole Centrale de Nantes,
                        Universit\'e de Nantes, \'Ecole des Mines de Nantes} \\
      1, rue de la No\"e, 44321 Nantes, France \\
      Damien.Chablat@irccyn.ec-nantes.fr \\
      Philippe.Wenger@irccyn.ec-nantes.fr \\
      Stephane.Caro@irccyn.ec-nantes.fr
    }
}
\author{J. Angeles
    \affiliation{
      Department of Mechanical Engineering \& \\
      Centre for Intelligent Machines, McGill University \\
      817 Sherbrooke Street West, Montreal, Canada H3A 2K6 \\
      angeles@cim.mcgill.ca
    }
}
\begin{document}
\maketitle
\begin{abstract}
The subject of this paper is a special class of three-degree-of-freedom parallel manipulators. The singular configurations of the two Jacobian matrices are first studied. The isotropic configurations are then found based on the characteristic length of this manipulator. The isoconditioning loci for the Jacobian matrices are plotted to define a global performance index allowing the comparison of the different working modes. The index thus resulting is compared with the Cartesian workspace surface and the average of the condition number.
\end{abstract}
\section{Introduction}
Various performance indices have been devised to assess the
kinetostatic performances of serial and parallel manipulators. The
literature on performance indices is extremely rich to fit in the
limits of this paper, the interested reader being invited to look
at it in the references cited here. A dimensionless quality index
was recently introduced in \cite{Lee} based on the ratio of the
Jacobian determinant to its maximum absolute value, as applicable
to parallel manipulators. This index does not take into account
the location of the operation point of the end-effector, because
the Jacobian determinant is independent of this location. The
proof of the foregoing result is available in \cite{Angeles97}, as
pertaining to serial manipulators, its extension to their parallel
counterparts being straightforward. The {\em condition number} of
a given matrix, on the other hand, is well known to provide a
measure of invertibility of the matrix \cite{Golub}. It is thus
natural that this concept found its way in this context. Indeed,
the condition number of the Jacobian matrix was proposed in
\cite{Salisbury} as a figure of merit to minimize when designing
manipulators for maximum accuracy. In fact, the condition number
gives, for a square matrix, a measure of the relative
roundoff-error amplification of the computed results  \cite{Golub}
with respect to the data roundoff error. As is well known,
however, the dimensional inhomogeneity of the entries of the
Jacobian matrix prevents the straightforward application of the
condition number as a measure of Jacobian invertibility. The {\em
characteristic length} was introduced in \cite{Angeles92} to cope
with the above-mentioned inhomogeneity.
\par
In this paper we use the {\em characteristic length} to normalize
the Jacobian matrix of a three-dof planar manipulator and to
calculate the isoconditioning loci for all its working modes.
\section{Preliminaries}
A planar three-dof manipulator with three parallel PRR chains, the
object of this paper, is shown in Fig.~1. This manipulator has
been frequently studied, in particular in
\cite{Merlet,Gosselin92}. The actuated joint variables are the
displacements of the three prismatic joints, the Cartesian
variables being the position vector \negr p of the operation point
$P$ and the orientation $\theta$ of the platform.

 \begin{figure}[ht]
  \begin{center}
    \includegraphics[width=80mm,height=66mm]{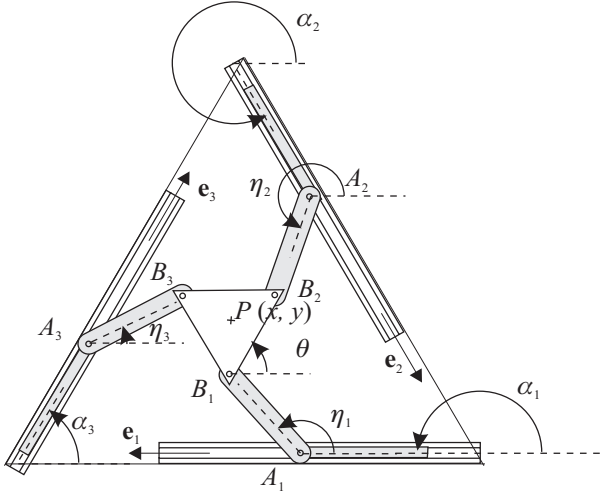}
    \caption{A three-DOF parallel manipulator}
    \protect\label{figure1}
  \end{center}
 \end{figure}
The trajectories of the points $A_i$ define an equilateral
triangle whose geometric center is the point $O$, while the points
$B_1$, $B_2$ and $B_3$, whose geometric center is the point $P$,
lie at the corners of an equilateral triangle. We thus have
$\alpha_i=\pi + (i-1)(2\pi/3)$, for $i=1,2,3$. Moreover,
$l=l_1=l_2=l_3$, with $l_i$ denoting the length of $A_iB_i$ and
$r=r_1=r_2=r_3$, with $r_i$ denoting the length of $B_i P$, in
units of length that need not be specified in the paper. The
layout of the trajectories of points $A_i$ is defined by the
radius $R$ of the circle inscribed in the associated triangle.
\subsection{Kinematic Relations}
The velocity $\dot{\bf p}$ of  point $P$ can be obtained in three
different forms, depending on which leg is traversed, namely,
 \bseq
  \beqa
  \dot{\negr p} = \dot{\negr a}_1
               +  \dot{\eta}_1 \negr E (\negr b_1 - \negr a_1)
               +  \dot{\theta} \negr E (\negr p - \negr b_1) \\
  \dot{\negr p} = \dot{\negr a}_2
               +  \dot{\eta}_2 \negr E (\negr b_2 - \negr a_2)
               +  \dot{\theta} \negr E (\negr p - \negr b_2) \\
  \dot{\negr p} = \dot{\negr a}_3
               +  \dot{\eta}_3 \negr E (\negr b_3 - \negr a_3)
               +  \dot{\theta} \negr E (\negr p - \negr b_3)
 \eeqa
 \label{e_1}
 \eseq
with matrix $\negr E$ defined as
 \bed
 {\bf E}= \left[\begin{array}{cr}
              0 & ~-1 \\
              1 &  0
             \end{array}
        \right]
 \eed
The velocity $\dot{\negr a_i}$ of points $A_i$ is given by
 \bed
  \dot{\negr a}_i= \dot{\rho}_i \frac{{\pmb \rho}_i}{||{ \pmb \rho}_i||}=
  \dot{\rho}_i \left[
  \begin{array}{c}
    \cos(\alpha_i) \\
    \sin(\alpha_i)
  \end{array}
  \right]= \dot{\rho}_i \negr e_i
 \eed
where $\negr e_i$ is a unit vector in the direction of the $i$th
prismatic joint.

We would like to eliminate the three idle joint rates
$\dot{\eta}_1$, $\dot{\eta}_2$ and $\dot{\eta}_3$ from
eqs.(\ref{e_1}a-c), which we do upon dot-multiplying the former by
$(\negr b_i - \negr a_i)^T$, thus obtaining
 \bseq
 \beqa
  (\negr b_1 - \negr a_1)^T \dot{\negr p} =
  (\negr b_1 - \negr a_1)^T \dot{\rho}_1 \negr e_1 +
  (\negr b_1 - \negr a_1)^T \dot{\theta} \negr E (\negr p - \negr b_1) \\
  (\negr b_2 - \negr a_2)^T \dot{\negr p} =
  (\negr b_2 - \negr a_2)^T \dot{\rho}_2 \negr e_2 +
  (\negr b_2 - \negr a_2)^T \dot{\theta} \negr E (\negr p - \negr b_2) \\
  (\negr b_3 - \negr a_3)^T \dot{\negr p} =
  (\negr b_3 - \negr a_3)^T \dot{\rho}_3 \negr e_3 +
  (\negr b_3 - \negr a_3)^T \dot{\theta} \negr E (\negr p - \negr b_3)
 \eeqa
 \label{e_2}
 \eseq
Equations (\ref{e_2}a-c) can now be cast in vector form, namely,
 \be
  {\bf A} {\negr t}={\bf B \dot{\pmb\rho}} \quad {\rm with} \quad
  \negr t=\left[\begin{array}{c}
                 \dot{\negr p} \\
                 \dot{\theta}
            \end{array}
          \right] \quad {\rm and } \quad
  \dot{\pmb \rho}=\left[\begin{array}{c}
                 \dot{\rho}_1 \\
                 \dot{\rho}_2 \\
                 \dot{\rho}_3
            \end{array}
          \right]
  \label{e:Adp=Bdth}
 \ee
with $\dot{\pmb{\rho}}$ thus being the vector of actuated joint
rates.

Moreover, \negr A and \negr B are, respectively, the
direct-kinematics and the inverse-kinematics matrices of the
manipulator, defined as
 \bseq
 \beqa
 \negr A&=& \left[\begin{array}{cc}
   (\negr b_1 - \negr a_1)^T & ~~
   -(\negr b_1 - \negr a_1)^T \negr E (\negr p - \negr b_1) \\
   (\negr b_2 - \negr a_2)^T & ~~
   -(\negr b_2 - \negr a_2)^T \negr E (\negr p - \negr b_2) \\
   (\negr b_3 - \negr a_3)^T & ~~
   -(\negr b_3 - \negr a_3)^T \negr E (\negr p - \negr b_3)
            \end{array}
         \right] \\
  \negr B&=&
  \left[\begin{array}{ccc}
   (\negr b_1 - \negr a_1)^T \negr e_1 &
   0 &
   0 \\
   0 &
   (\negr b_2 - \negr a_2)^T \negr e_2 &
   0 \\
   0 &
   0 &
   (\negr b_3 - \negr a_3)^T \negr e_3
        \end{array}
  \right]
 \label{equation:matrices_B}
 \eeqa
 \eseq
When \negr A and \negr B are nonsingular, we obtain the relations
 \bed
   \negr t = \negr J \dot{\pmb \rho} {\rm,~~with~~} \negr J = \negr A^{-1} \negr B
   \quad {\rm and} \quad
   \dot{\pmb \rho} = \negr K \negr t
 \eed
with \negr K denoting the inverse of \negr J.
\subsection{Parallel Singularities}
\begin{figure}[hbt]
    \begin{center}
    \begin{tabular}{ccc}
       \begin{minipage}[t]{45 mm}
    \includegraphics[width=42mm,height=37mm]{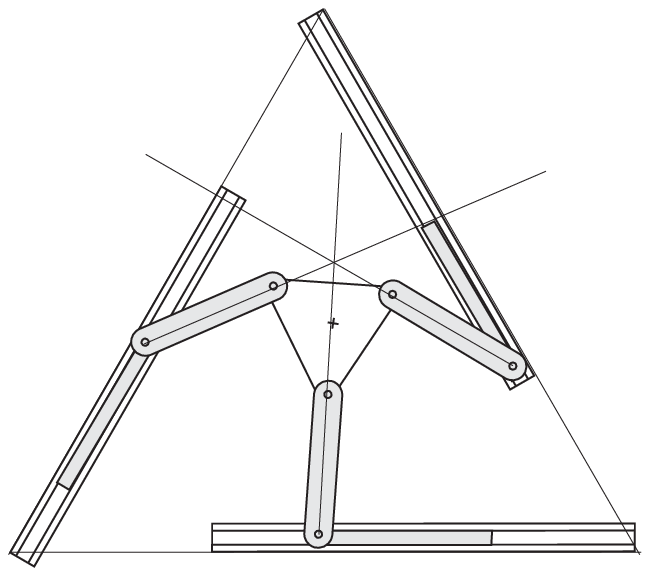}
    \caption{Parallel singularity}
    \protect\label{figure2}       \end{minipage} &
       \begin{minipage}[t]{45 mm}
    \includegraphics[width=42mm,height=37mm]{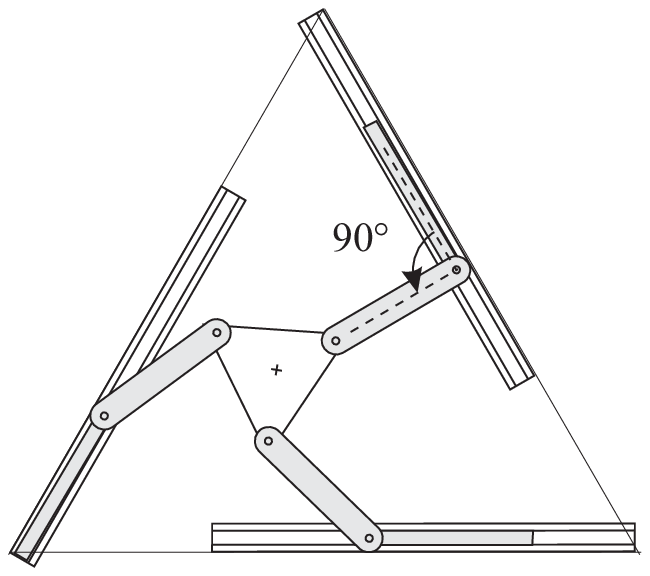}
    \caption{Serial singularity}
    \protect\label{figure3}
       \end{minipage}
    \end{tabular}
    \end{center}
\end{figure}

Parallel singularities occur when the determinant of matrix \negr
A vanishes \cite{Chablat,Gosselin90}. At these configurations, it
is possible to move locally the operation point $P$ with the
actuators locked, the structure thus resulting cannot resist
arbitrary forces, and control is lost. To avoid any performance
deterioration, it is necessary to have a Cartesian workspace free
of parallel singularities. For the planar manipulator studied,
such configurations are reached whenever the axes $A_1B_1$,
$A_2B_2$ and $A_3B_3$ intersect (possibly at infinity), as
depicted in Fig.~\ref{figure2}.

In the presence of such configurations, moreover, the manipulator
cannot resist a force applied at the intersection point. These
configurations are located inside the Cartesian workspace and form
the boundaries of the joint workspace \cite{Chablat}.
\subsection{Serial Singularities}
Serial singularities occur when $\det(\negr B) = 0$. In the
presence of theses singularities, there is a direction along which
no Cartesian velocity can be produced. Serial singularities define
the boundary of the Cartesian workspace. For the topology under
study, the serial singularities occur whenever $(\negr b_i - \negr
a_i)^T \negr e_i=0$ for at least one value of $i$, as depicted in
Fig.~\ref{figure3} for $i=2$.
\section{Isoconditioning Loci}
\subsection{The Matrix Condition Number}
We derive below the loci of equal condition number of the matrices
\negr A, \negr B and \negr K. To do this, we first recall the
definition of condition number of an $m \times n$ matrix $\negr
M$, with $m \leq n$, $\kappa(\negr M)$. This number can be defined
in various ways; for our purposes, we define $\kappa(\negr M)$ as
the ratio of the smallest, $\sigma_s$, to the largest, $\sigma_l$,
singular values of $\negr M$, namely,
 \be
   \kappa(\negr M) =  \frac{\sigma_s}{\sigma_l}
 \ee
The singular values $\left\{\sigma_i\right\}_1^m$ of matrix $\negr
M$ are defined, in turn, as the square roots of the nonnegative
eigenvalues of the positive-definite $m \times m$ matrix $\negr M
\negr M^T$.
\subsection{Non-Homogeneous Direct-Kinematics Matrix}
To render the matrix $\negr A$ homogeneous, as needed to define
its condition number, each term of the third column of $\negr A$
is divided by the characteristic length $L$ \cite{Ranjbaran},
thereby deriving its normalized counterpart $\overline{\negr A}$:
 \be
  \overline{\negr A}=
  \left[\begin{array}{cc}
        (\negr b_1-\negr a_1)^T & ~~-(\negr b_1-\negr a_1)^T \negr E (\negr p - \negr b_1)/L \\
        (\negr b_2-\negr a_2)^T & ~~-(\negr b_2-\negr a_2)^T \negr E (\negr p - \negr b_2)/L \\
        (\negr b_3-\negr a_3)^T & ~~-(\negr b_3-\negr a_3)^T \negr E (\negr p - \negr b_3)/L
        \end{array}
  \right]
 \ee
which is calculated so as to minimize $\kappa(\overline{\negr
A})$, along with the posture variables $\rho_1$, $\rho_2$ and
$\rho_3$.

However, notice that \negr B is dimensionally homogeneous, and
need not be normalized.
\subsection{Isotropic Configuration}
In this section, we derive the isotropy condition on \negr J to
define the geometric parameters of the manipulator. We shall
obtain also the value $L$ of the characteristic length. To
simplify $\overline{\negr A}$ and \negr B, we use the notation
 \bseq
 \beqa
   \negr l_i&=& (\negr b_i - \negr a_i )   \\
   k_i      &=& (\negr b_i - \negr a_i)^T \negr E (\negr p - \negr b_i)    \\
   m_i      &=& (\negr b_i - \negr a_i)^T \negr e_i   \\
   \gamma_i      &=& \angle A_i B_i P
 \eeqa
 \label{e_notation}
 \eseq
We can thus write matrices $\overline{\negr A}$ and $\negr B$ as
 \beqa
  \overline{\negr A}=
  \left[\begin{array}{cc}
                 \negr l_1^T & -k_1 / L\\
                 \negr l_2^T & -k_2 / L\\
                 \negr l_3^T & -k_3 / L
        \end{array}
  \right] \quad
  \negr B=
  \left[\begin{array}{ccc}
            m_1 & 0   & 0 \\
            0   & m_2 & 0 \\
            0   & 0   & m_3
        \end{array}
  \right]
 \eeqa
Whenever matrix \negr B is nonsingular, that is, when $m_i \neq
0$, for $i=1,2,3$, we have
 \bed
   \overline{\negr K}=
  \left[\begin{array}{ccc}
     \negr l_1^T/m_1  &
     ~~-k_1/(L~ m_1) \\
     \negr l_2^T/m_2  &
     ~~-k_2/(L~ m_1) \\
     \negr l_3^T/m_3  &
     ~~-k_3/(L~ m_3)
        \end{array}
  \right]
 \eed
Matrix $\overline{\negr J}$, the normalized \negr J, is isotropic
if and only if $\overline{\negr K} \overline{\negr K}^{T}= \tau^2~
\negr 1_{3 \times 3}$ for $\tau> 0$ and $\overline{\negr
K}=\overline{\negr J}^{-1}$, {\it i.e.},
 \bseq
  \beqa
   (\negr l_1^T \negr l_1+k_1^2/L^2)/m_1^2= \tau^2  \\
   (\negr l_2^T \negr l_2+k_2^2/L^2)/m_2^2= \tau^2  \\
   (\negr l_3^T \negr l_3+k_3^2/L^2)/m_3^2= \tau^2  \\
   (\negr l_1^T \negr l_2+k_1 k_2/L^2)/(m_1 m_2)=0   \\
   (\negr l_1^T \negr l_3+k_1 k_3/L^2)/(m_1 m_3)=0   \\
   (\negr l_2^T \negr l_3+k_2 k_3/L^2)/(m_2 m_3)=0
  \eeqa
  \label{e_isotropy}
 \eseq
From eqs.(\ref{e_isotropy}a-f), we can derive the conditions
below:
 \bseq
  \beqa
   ||\negr l_1|| = ||\negr l_2|| = ||\negr l_3||   \\
   ||\negr p-\negr b_1|| = ||\negr p-\negr b_2|| = ||\negr p-\negr b_3||   \\
   \negr l_1^T \negr l_2 = \negr l_2^T \negr l_3 = \negr l_2^T \negr l_3   \\
   m_1 m_2 = m_1 m_3 = m_2 m_3
  \eeqa
  \label{e_isotropy_cond}
 \eseq
In summary, the constraints defined in the
eqs.(\ref{e_isotropy_cond}a-d) are:
\begin{description}
   \item[$\circ$] Pivots $B_i$ should be placed at the vertices of an equilateral
triangle;
   \item[$\circ$] Segments $A_iB_i$ form an equilateral triangle;
   \item[$\circ$] The trajectories of point $A_i$ define an equilateral triangle an hence,
$l=l_1= l_2=l_3$.
\end{description}
Notice that the foregoing conditions, except the second one, were
assumed in $\S 2$.
\subsection{The Characteristic Length}
The characteristic length is defined at the isotropic
configuration. From eqs.(\ref{e_isotropy}d-f), we determine the
value of the characteristic length as,
 \bed
   L = \sqrt{\frac{-k_1 k_2}{\negr l_1^T \negr l_2}}
 \eed
By applying the constraints defined in eqs.(\ref{e_isotropy}a-d),
we can write the characteristic length $L$ in terms of angle
$\gamma$, {\it i.e.}
 \bed
   L= \sqrt{2} r \sin(\gamma)
 \eed
where $\gamma=\gamma_1=\gamma_2=\gamma_3$, was defined in
eq.(\ref{e_notation}d) and $\gamma \in [0~2\pi]$.

This means that the manipulator under study admits several
isotropic configurations, two of which are shown in Figs.~4a and
b, whereas the characteristic length $L$ of a manipulator is
unique \cite{Angeles97}. When $\gamma$ is equal to $\pi/2$, Fig.
4a, {\it i.e.}, when $A_iB_i$ is perpendicular to $B_iP$, the
manipulator finds itself at a configuration furthest away from
parallel singularities. To have an isotropic configuration
furthest away from serial singularities, we have two conditions:
$\negr e_i^T \negr E (\negr b_i - \negr a_i)=0$ and $r=R/2$.
 \begin{figure}[hbt]
  \begin{center}
    \includegraphics[width=86mm,height=40mm]{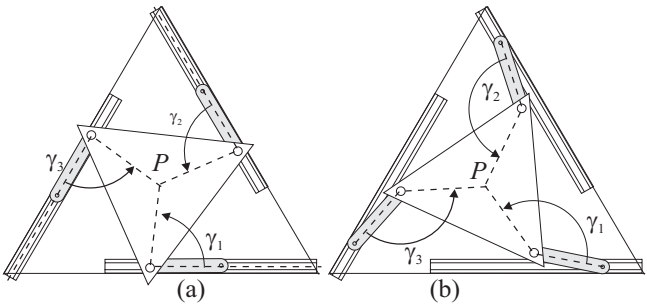}
    \caption{Two isotropic configurations with two values of $\rho$}
    \protect\label{figure4}
  \end{center}
 \end{figure}
\subsection{Working Modes}
The manipulator under study has a diagonal inverse-kinematics
matrix \negr B, as shown in eq.(5b), the vanishing of one of its
diagonal entries thus indicating the occurrence of a serial
singularity. The set of manipulator postures free of this kind of
singularity is termed a working mode. The different working modes
are thus separated by a serial singularity, with a set of postures
in different working modes corresponding to an inverse kinematics
solution.

The formal definition of the working mode is detailed in
\cite{Chablat}. For the manipulator at hand, there are eight
working modes, as depicted in Fig. 5.

 \begin{figure}[hbt]
  \begin{center}
    \includegraphics[width=87mm,height=42mm]{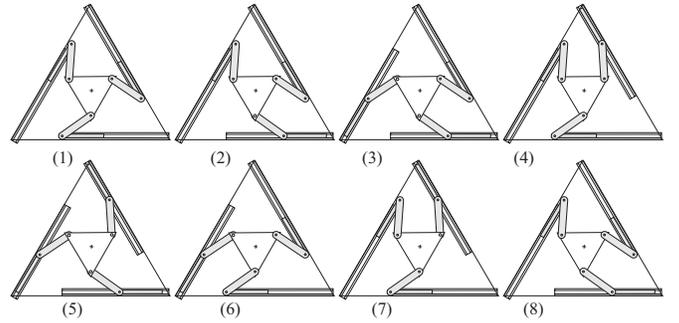}
    \caption{The eight working modes of the $3PRR$ manipulator}
    \protect\label{figure5}
  \end{center}
 \end{figure}
However, because of symmetries, we can restrict our study to only
two working modes, if there are no joint limits. Indeed, working
mode $1$ is similar to working mode $5$, because for the first
one, the signs of the diagonal entries of $\negr B$ are all
negative, and for the second are all positive. A similary
reasoning is applicable to the working modes $2$-$6$, $3$-$7$ and
$4$-$8$; likewise, the working modes $3$-$4$ and $7$-$8$ can be
derived from the working modes 2 and 6 by a rotation of
$120^{\circ}$ and $240^{\circ}$, respectively. Therefore, only the
working modes $1$ and $2$ are studied. We label the corresponding
matrices as $\overline{\negr A}_i$, $\negr B_i$, $\overline{\negr
K}_i$ for the $i$th working mode.
\subsection{Isoconditioning Loci}
For each Jacobian matrix and for all the poses of the
end-effector, we calculate the optimum conditioning according to
the orientation of the end-effector. We can notice that for any
orientation of the end-effector, there is a singular
configuration.

\begin{figure}[hbt]
  \begin{center}
    \includegraphics[width=90mm,height=44mm]{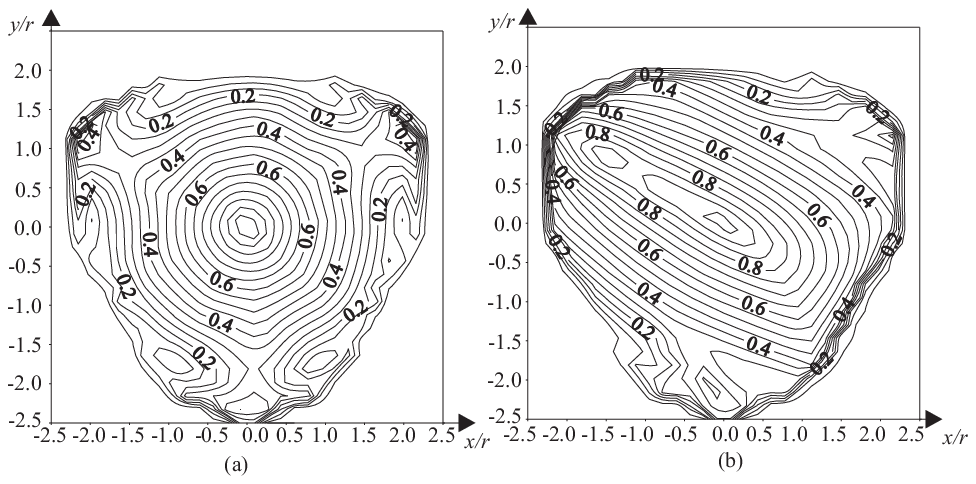}
    \caption{Isoconditioning loci of the matrix
    (a) $\overline{\negr A}_1$ and (b) $\overline{\negr A}_2$ with $R/r=2$ and $l/r=2$}
    \protect\label{figure6}
  \end{center}
 \end{figure}

Figure~~\ref{figure6} depicts the isoconditioning loci of matrix
$\overline{\negr A}$.

\begin{figure}[hbt]
  \begin{center}
    \includegraphics[width=90mm,height=40mm]{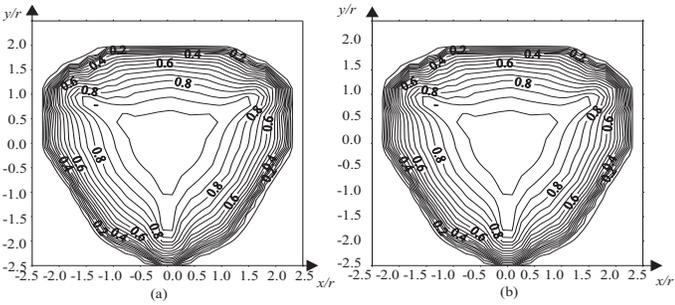}
    \caption{Isoconditioning loci of the matrix
    (a) $\negr B_1$ and (b) $\negr B_2$  with $R/r=2$ and $l/r=2$}
    \protect\label{figure7}
  \end{center}
 \end{figure}

We depict in Fig.~\ref{figure7} the isoconditioning loci of matrix
\negr B. We notice that the loci of both working modes are
identical. This is due to both the absence of joint limits on the
actuated joints and the symmetry of the manipulator. For one
configuration, only the signs of $m_i$ change from a working mode
to another, but the condition number $\kappa$ is computed from the
absolute values of $m_i$.

The shapes of the isoconditioning loci of $\overline{\negr K}$
(Fig. 8) are similar to those of the isconditioning loci of
$\overline{\negr A}$; only the numerical values change.
 \begin{figure}[hbt]
  \begin{center}
    \includegraphics[width=88mm,height=42mm]{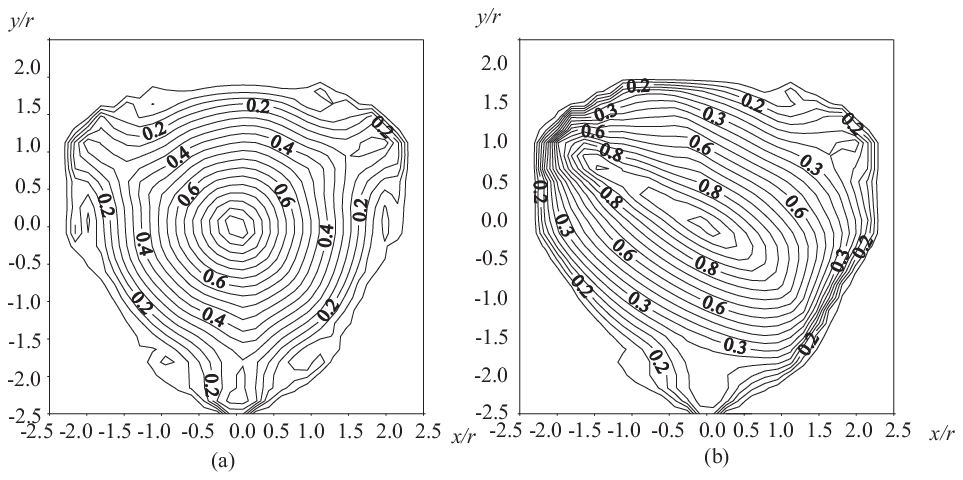}
    \caption{Isoconditioning loci of the matrix
    (a) $\overline{\negr K}_1$ and (b) $\overline{\negr K}_2$ with $R/r=2$ and $l/r=2$}
    \protect\label{figure8}
  \end{center}
 \end{figure}

For the first working mode, the condition number  of both matrices
$\overline{\negr A}$ and $\overline{\negr K}$ decreases regularly
around the isotropic configuration. The isoconditioning loci
resemble concentric circles. However, for the second working mode,
the isoconditioning loci of both matrices $\overline{\negr A}$ and
$\overline{\negr K}$ resemble ellipses.

The characteristic length $L$ depends on $r$. Two indices can be
studied according to parameter $R$: (i) the area of the Cartesian
workspace, called $\cal S$, and (ii) the average of the
conditioning, called $\overline{\kappa}$.

\begin{figure}[!hb]
  \begin{center}
    \includegraphics[width=61mm,height=27mm]{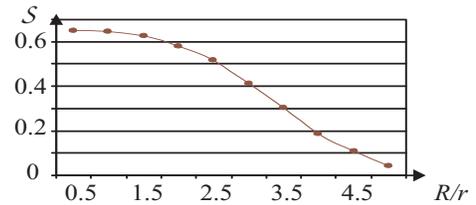}
    \caption{Variation of $\cal S$ as a function of $R/r$, with $l/r=2$}
    \protect\label{workspace_surface}
  \end{center}
 \end{figure}

The first index is identical for the two working modes.
Figure~\ref{workspace_surface} depicts the variation of $S$ as a
function of $R/r$, for $l/r=2$. Its maximum value is reached when
$R/r=0.5$.

\begin{figure}[!ht]
  \begin{center}
    \includegraphics[width=69mm,height=46mm]{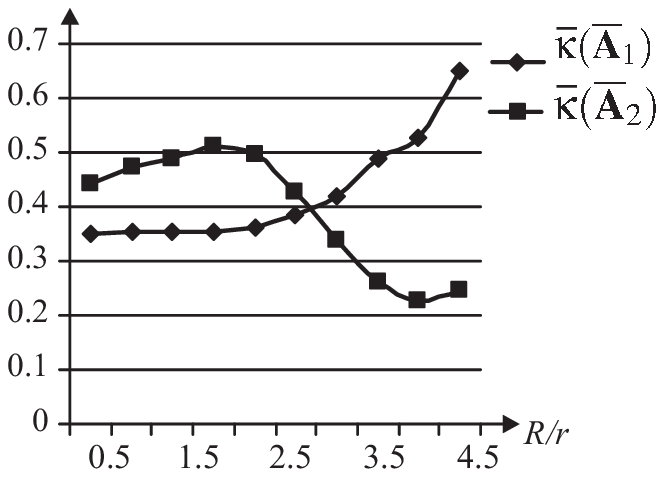}
    \caption{$\overline{\kappa}(\overline{\negr A}_1)$ and $\overline{\kappa}(\overline{\negr A}_2)$ as a function of $R/r$, with $l/r=2$ }
    \protect\label{figure12}
  \end{center}
 \end{figure}

\begin{figure}[!ht]
  \begin{center}
    \includegraphics[width=66mm,height=28mm]{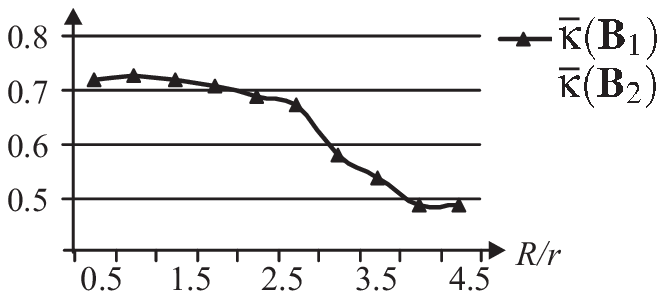}
    \caption{$\overline{\kappa}(\negr B_1)$ and $\overline{\kappa}(\negr B_2)$ as a function of $R/r$, with $l/r=2$}
    \protect\label{figure13}
  \end{center}
 \end{figure}

\begin{figure}[!ht]
  \begin{center}
    \includegraphics[width=73mm,height=35mm]{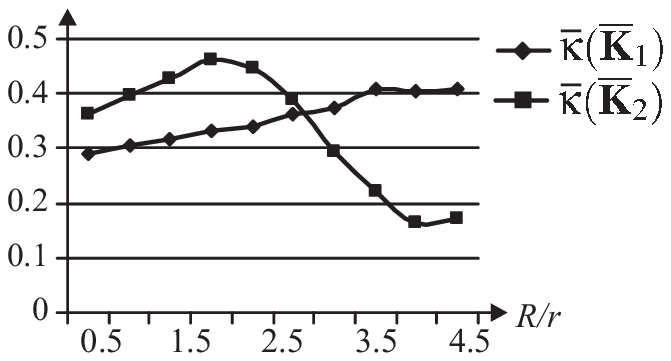}
    \caption{$\overline{\kappa}(\overline{\negr K}_1)$ and $\overline{\kappa}(\overline{\negr K}_2)$ as a function of $R/r$, with $l/r=2$}
    \protect\label{figure14}
  \end{center}
 \end{figure}

For the three matrices studied, $\overline{\kappa}$ can be
regarded as a global performance index. This index thus allows us
to compare the working modes. Figsures~\ref{figure12},
\ref{figure13} and \ref{figure14} depict
$\overline{\kappa}(\overline{\negr A})$, $\overline{\kappa}(\negr
B)$ and $\overline{\kappa}(\overline{\negr K})$, respectively, as
a function of $R/r$, with $l/r=2$.

The value of $\overline{\kappa}(\overline{\negr A}_1)$ increases
with $R$. At the opposite,  the maximum value of
$\overline{\kappa}(\overline{\negr A}_2)$ is reached when $R/r=2$.
For both the working modes studied, $\overline{\kappa}(\negr B_1)$
and $\overline{\kappa}(\negr B_2)$ are identical for $R/r$ fixed.
For the first working mode, the minimum value of
$\overline{\kappa}(\overline{\negr K}_1)$ and the maximum area of
Cartesian workspace $\cal S$ occur at different values of $R/r$.
This means that we must reach a compromise under these two
indices. For the second working mode, there is an optimum of
$\overline{\kappa}(\overline{\negr K}_2)$ close to the optimum of
$\cal S$, for $R/r=2$.
\section{Conclusions}
We produced the isoconditioning loci of the Jacobian matrices of a
three-PRR parallel manipulator. This concept being general, it can
be applied to any three-dof planar parallel manipulator. To solve
the problem of nonhomogeneity of the Jacobian matrix, we used the
notion of characteristic length. This length was defined for the
isotropic configuration of the manipulator. The isoconditioning
curves thus obtained characterize, for every posture of the
manipulator, the optimum conditioning for all possible
orientations of the end-effector. This index is compared with the
area of Cartesian workspace and the conditioning average. The two
optima being different, it is necessary to find another index to
determine the optimum values. The results of this paper can be
used to choose the working mode which is best suited to the task
at hand or as a global performance index when we study the optimum
design of this kind of manipulators.
\bibliographystyle{unsrt}

\end{document}